\documentclass[letterpaper]{article} 
\usepackage{aaai25}  
\usepackage{times}  
\usepackage{helvet}  
\usepackage{courier}  
\usepackage[hyphens]{url}  
\usepackage{graphicx} 
\usepackage{natbib}  
\usepackage{caption} 
\usepackage{algorithm}
\usepackage{algorithmic}

\usepackage{newfloat}
\usepackage{listings}
\DeclareCaptionStyle{ruled}{labelfont=normalfont,labelsep=colon,strut=off} 
\floatstyle{ruled}
\newfloat{listing}{tb}{lst}{}
\floatname{listing}{Listing}
\usepackage{times}
\usepackage{latexsym}
\usepackage{multirow}
\usepackage{dsfont}
\usepackage{amsmath, amsthm, mathtools, amssymb}
\usepackage{svg}
\usepackage{amsfonts}
\usepackage[shortlabels, inline]{enumitem}

\definecolor{BrickRed}{HTML}{B6321C}
\definecolor{ForestGreen}{HTML}{009B55}
\definecolor{Mulberry}{HTML}{A93C93}

\title{MultiQ\&A: An Analysis in Measuring Robustness\\via Automated Crowdsourcing of Question Perturbations and Answers}
\author{
    Nicole Cho\equalcontrib,
    William Watson\equalcontrib
}
\affiliations{
    J.P. Morgan AI Research\\

    383 Madison Ave.\\
    New York, NY USA\\
    nicole.cho@jpmorgan.com
}

\usepackage{bibentry}

\begin{document}

\maketitle

\begin{abstract}
One critical challenge in the institutional adoption journey of Large Language Models (LLMs) stems from their propensity to hallucinate in generated responses. To address this, we propose MultiQ\&A, a systematic approach for evaluating the robustness and consistency of LLM-generated answers. We demonstrate MultiQ\&A's ability to crowdsource question perturbations and their respective answers through independent LLM agents at scale. Our experiments culminated in the examination of 1.9 million question perturbations and 2.3 million answers. Furthermore, MultiQ\&A shows that ensembled LLMs, such as \texttt{gpt-3.5-turbo}, remain relatively robust and consistent under perturbations. MultiQ\&A provides clarity in the response generation space, offering an effective method for inspecting disagreements and variability. Therefore, our system offers a potential framework for institutional LLM adoption with the ability to measure confidence, consistency, and the quantification of hallucinations.
\end{abstract}

%

\begin{figure*}[!ht]
\centering
\includegraphics[clip, trim=0cm 3.85cm 0cm 0.3cm,width=0.9\textwidth]{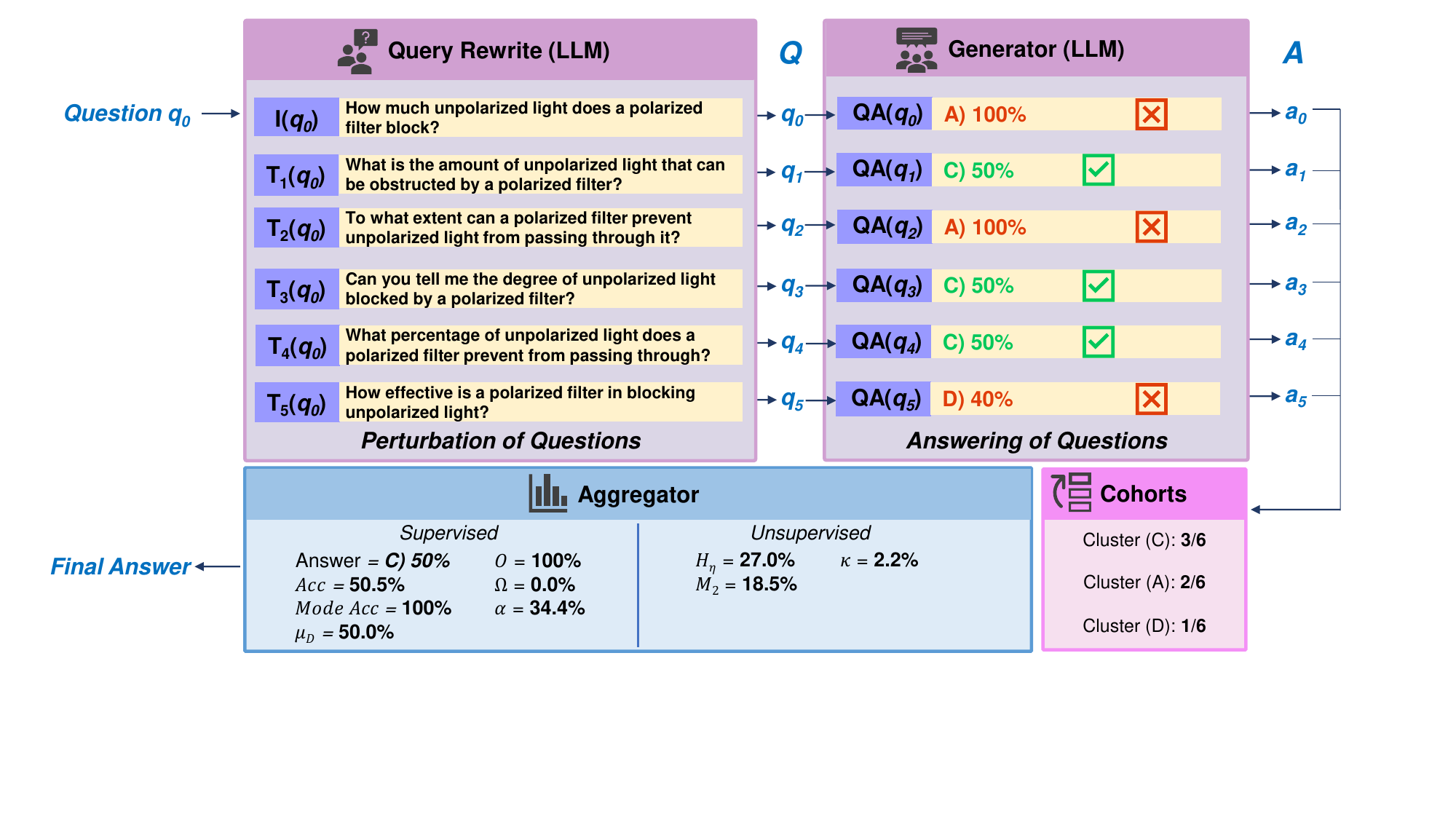}
\caption{System Overview for MultiQ\&A: A single question $q_0$, supplied by the user, is perturbed in $v$ different ways (while retaining the original question via the identify function). Each perturbed question $q_i\in\mathcal{T}$ is independently answered by the Answer Generator agent. Finally, several metrics are computed for the cohort of answers based on the perturbations. In a practical setting, these variations can be fed into an Aggregator, which organizes and re-ranks the answers according to the user's preferences and the original question. Aggregated statistics are compiled from $1,000$ random permutations of the result set across raters, with labels remapped, thus simulating large-scale item analysis.}
\label{fig:system_overview}
\vspace{-3mm}
\end{figure*}


\section{Introduction}

Large Language Models (LLMs) exhibit immense potential for diverse downstream applications; however, they often lack transparency regarding their reasoning processes \citep{liang2022holistic, wei2023chainofthought, kojima2023large, li2023making} and the robustness of their generated answers. This challenge is further exacerbated by the limited accessibility into a models' training datasets, especially when deployed externally \citep{liang2022holistic}. Moreover, research has demonstrated that LLMs are highly sensitive to input perturbations \citep{zhang2022interpretingrobustnessneuralnlp, moradi2021evaluatingrobustnessneurallanguage}. Therefore, the motivation for this study is to propose an evaluation method for LLMs that is undeterred by such limitations and operates on the external interface. We present MultiQ\&A, an adversarial ``IQ test" for language models, mainly \texttt{gpt-3.5-turbo}, that measures the robustness of answer generation by automating the crowdsourcing of questions and answers through independent agents. This cognitive evaluation method stands in contrast to the more commonly utilized context-based retrieval systems for hallucination mitigation \citep{reimers-2019-sentence-bert, johnson2019billion, nogueira2020passage, karpukhin-etal-2020-dense, NEURIPS2020_6b493230, izacard-grave-2021-leveraging}. 
Other methods include relying on the model's general knowledge \citep{khashabi-etal-2020-unifiedqa} or conditioning the QA model on context generated by the LLM itself \citep{yu2023generate}. In comparison,  MultiQ\&A employs a five-pronged approach that brings forth four main contributions. Our approach consists of:







\begin{enumerate}[noitemsep, leftmargin=*, topsep=0pt, partopsep=0pt]
\item Stress-test 365,000 questions by perturbing into diverse lexical variants while retaining the original semantics.
\item Crowdsource answers using independent LLM agents under different perturbations.
\item Quantify response diversity and measure robustness.
\item Ensemble answers through plurality voting.
\item Visualize disagreements to identify hallucinations.
\end{enumerate}
As a result, our study culminates in the following four main pillars of contribution: 
\begin{itemize}[noitemsep, leftmargin=*, topsep=0pt, partopsep=0pt, label={\tiny\raisebox{0.5ex}{$\blacktriangleright$}}]
    \item Analyzed over 1.9 million perturbed questions and 2.3 million answers across extractive, open-ended, and multiple-choice QA, mimicking real-life scenarios.
    
    \item Demonstrated the capacity of \texttt{gpt-3.5-turbo} to generate semantically stable yet lexically diverse transformations, providing insight into its reasoning capabilities. MultiQ\&A's adversarial game automatically generates large sets of questions and highlights the model's variability for each question, as shown in Figure \ref{fig:system_overview}.
    
    \item Therefore, we introduce MultiQ\&A as a robust and scalable framework to stress-test LLMs in its ability to perform under perturbed questions. In this context, MultiQ\&A likens itself to an ``IQ Test" or a cognitive evaluation for language models.  
    
    \item Finally, MultiQ\&A enables granular analysis of question-answer pairs to reveal subtle inconsistencies and areas of hallucination, guiding future model improvements.

\end{itemize}
Additionally, our methodology can offer insights into alternative tasks related to question-answering, such as assessing the differences between ambiguous and consistent queries, understanding different inputs that trigger content-filters, and enumerating re-phrasings that are adversarial.

\section{Related Work}
Large language models (LLMs), such as GPT-3, InstructGPT, and LLaMA \citep{NEURIPS2020_1457c0d6, ouyang2022training, touvron2023llama}, have demonstrated remarkable capabilities but also face challenges, such as hallucinations and sensitivity to input perturbations. To mitigate hallucinations, techniques like chain-of-thought \citep{wei2023chainofthought} and step-by-step generation \citep{nye2021work} have been proposed. Other strategies include augmenting generation with semantic retrieval \citep{liu2021makes,reimers-2019-sentence-bert}, generating context directly \citep{yu2023generate}, and crafting multi-step chains using tools like PromptChainer \citep{10.1145/3491101.3519729} or probabilistic programs \citep{dohan2022language}. Self-consistency \citep{wang2023selfconsistency} and gradient-based sampling \citep{kumar-etal-2022-gradient} further improve reliability by refining decoding processes. Modular architectures in legal and financial domains demonstrate how domain-specific tools and workflows can overcome context limitations and achieve competitive performance \citep{watson-etal-2025-law, 10.1145/3677052.3698597, 10.1145/3604237.3626908, watson-etal-2023-hiddentables}.

\paragraph{Perturbations in LLMs}
LLMs are highly sensitive to noisy inputs, which can impact real-world performance \citep{zhang2022interpretingrobustnessneuralnlp}, with existing benchmarks often failing to assess robustness accurately \citep{moradi2021evaluatingrobustnessneurallanguage}. \citet{slobodkin-etal-2023-curious} found that LLMs encode query answerability within their hidden states, suggesting potential avenues for decoding improvements. Similarly, \citet{azaria2023internalstatellmknows} demonstrated that LLMs' hidden states contain signals useful for detecting truthfulness. \citet{mallen-etal-2023-trust} highlighted the limitations of parametric memory in recalling long-tail knowledge and advocated for retrieval-augmented generation (RAG) systems to improve knowledge retention and efficiency.

\begin{table*}[!ht]
\centering
\small
\begin{tabular}{cl|lr}
\hline
\textbf{Domain}  & \textbf{Dataset}  & \textbf{Example Question} & \textbf{Answer} \\
\hline 
Falsehoods     & TruthfulQA &
Which is denser, water vapor or air? & Air\\ \hline
Physical Env.  & PIQA &
To cut a length of PVC to size, you can? & Use a saw\\ \hline
Yes/No         & BoolQ &
Are there 51 states in the United States? & No \\ \hline
STEM   & \multirow{3}{*}{MMLU} &
The strongest base in liquid ammonia is? & $\text{NH}_2^-$ \\
Humanities & &
When did the first pharaohs emerge in Egypt? & 5100 B.P.\\
Social Sciences &  &
The government measures inflation using? & CPI
\\ \hline
\multirow{4}{*}{Science} & OpenBookQA &
What raw material is consumed by chloroplast? & $\text{CO}_2$\\
& SciQ & 
Which is the final step of cell division? & Cytokinesis \\
& ARC - (C)  &
How many valence electrons does selenium have?	& $6$ \\
& ARC - (E)  &
Where is water most likely to be brackish? & Estuary\\ \hline
Mathematics & MathQA & 
If $n = 2^{0.15}$ and $n^b = 8$ , $b$ must equal? & $20$\\ \hline
\multirow{3}{*}{Wikipedia} & SQuADv2 &
Where is the Mona Lisa housed? & The Louvre \\
& WikiQA & 
What is korean money called? & The won\\
& HotpotQA & 
EMU and Ugg boots both originated from where? & Australia\\ \hline
General & TriviaQA & 
In an opera, whose lover was Cavaradossi? & Tosca\\
\hline
\end{tabular}
\caption{\label{tab:overview}
Overview of the 12 question-answering datasets studied in this work, the domain coverage, and examples of the question-answer format. These datasets span traditional QA formats such as {\color{BrickRed} \textbf{Extractive}}, {\color{ForestGreen} \textbf{Multiple Choice}}, and {\color{Mulberry} \textbf{Abstractive}}. Our experiments treat all scenarios as text generation tasks, albeit with different prompting templates to align responses with the ground truth answer.
}
\vspace{-3mm}
\end{table*}

\section{Methodology}
We propose a robust multi-step pipeline for measuring question answering (QA) robustness, consisting of three components: the Query Rewriter, Answer Generator, and Aggregator. These components (\texttt{gpt-3.5-turbo} agents) work in tandem to perturb, answer, and consolidate diverse responses, enabling automatic crowdsourcing and robust QA evaluation mimicking real-world scenarios.


\subsection{Query Rewriter}
The Query Rewriter, powered by \texttt{gpt-3.5-turbo}, transforms the original query $q_0$ into a diverse set of $v + 1$ variations $\mathcal{Q}=\{q_0, q_1, \dots, q_v\}$, ensuring semantic consistency while introducing meaningful perturbations. These variations include the identity transformation $T_0(q_0) = q_0$, ensuring that the original query is always preserved.

\paragraph{Perturbation Process} To encourage creativity and prevent duplicates, the perturbation process employs a high temperature setting ($\tau = 1.0$) in a single generation step. The resulting perturbations $\mathcal{Q}$ are generated as:
\begin{gather*}
\mathcal{Q} = \mathcal{T}(q_0) = \left[\begin{array}{c}
I(q_0) \\
T_1(q_0)  \\
\vdots \\
T_v(q_0) \end{array}\right] = \left[\begin{array}{c}
q_0 \\
q_1  \\ \vdots \\ 
q_v  \end{array}\right]
\end{gather*}
where $T_j$ represents perturbation functions in the transformation set $\mathcal{T}$. The output is a set of $v + 1$ semantically similar yet syntactically diverse questions, ready for answering.

\paragraph{Prompt Design} Each perturbed question is explicitly designed to preserve the semantic intent of $q_0$ while enabling evaluation of the system's robustness under diverse phrasings. A single prompt call generates all variations in one pass, leveraging the efficiency of LLMs to simulate creative crowdsourcing. The prompt is: \texttt{Rewrite the question in $n$ radically different ways}.

\subsection{Answer Generator}
The Answer Generator employs $|\mathcal{Q}| = v + 1$ independent \texttt{gpt-3.5-turbo} agents to generate answers $\mathcal{A} = \{a_0, a_1, \dots, a_v\}$ for each question $q_i \in \mathcal{Q}$. This design ensures no contextual information is shared amongst agents, isolating the effect of each query perturbation.

\paragraph{Answering Process}
For each query $q_0 \in \mathcal{Q}$, the agent receives a specific prompt, tailored to the type of QA task:
\begin{itemize}[noitemsep, leftmargin=*, topsep=0pt, partopsep=0pt, label={\tiny\raisebox{0.5ex}{$\blacktriangleright$}}]
    \item {\color{BrickRed} \textbf{Extractive QA}}: The question is presented with its corresponding context $c_i$, and the model extracts the answer.
    \item {\color{ForestGreen} \textbf{Multiple Choice QA}}: The question is presented alongside candidate choices $\mathcal{K} = \{k_0, \dots k_m\}$, with the model selecting the most appropriate option(s).
    \item {\color{Mulberry} \textbf{Abstractive QA}}: The question is presented in isolation, and the model generates a free-form answer.
\end{itemize}
The prompt formats for each scenario are detailed in Table~\ref{tab:format}. All experiments use a high temperature setting ($\tau = 1.0$) to prioritize diversity and stress-test the answering agents.

\subsection{Aggregator}
The Aggregator consolidates and evaluates the generated answers $\mathcal{A}$ using semantic clustering, ranking, and evaluation metrics. It provides holistic insights into the system's robustness and reliability through the following mechanisms:

\vspace{3pt}
\noindent
\textbf{Clustering} A semantic paraphrase model groups answers into coherent clusters based on similarity \citep{reimers-2019-sentence-bert}. This clustering provides an interpretable structure for analyzing cohort diversity and agreement.

\vspace{3pt}
\noindent
\textbf{Re-ranking} Within each cluster, an answer-critic model re-ranks responses to identify the most semantically aligned answer to the original query $q_0$. This alignment is measured using cross-encoder techniques via Sentence-BERT models \citep{reimers-2019-sentence-bert}.

\vspace{3pt}
\noindent
\textbf{Evaluation Metrics} The Aggregator computes both \textbf{supervised (S)} and \textbf{unsupervised (U)} metrics to quantify robustness and agreement (as defined in \S\ref{sec:metrics} and illustrated in Figure~\ref{fig:system_overview}).
Supervised metrics are computed when the ground truth is available, while unsupervised metrics analyze response consistency across perturbations. During live inference, users can select a cohort of answers as the ground truth, enabling real-time computation of supervised metrics. For exploratory analysis, unsupervised metrics are sufficient to assess LLM robustness under varying conditions.

\begin{table}[!t]
\centering
\small
\begin{tabular}{l|l}
\hline
\textbf{Scenario} & \textbf{Prompt Template} \\
\hline 
\multirow{3}{*}{\color{BrickRed} \textbf{Extractive}} & \texttt{Context: \{$c_i$\}} \\
& \texttt{Question: \{$q_i$\}} \\
& \texttt{Answer:} \\
\hline
\multirow{3}{*}{\color{ForestGreen} 
\begin{tabular}{@{}l} \textbf{Multiple} \\ \textbf{Choice} \end{tabular}
} 
& \texttt{Question: \{$q_i$\}} \\
& \texttt{A) \{$k_0$\}}\;\;\;\texttt{$\dots$}\;\;\;\texttt{Z) \{$k_m$\}} \\
& \texttt{Answer:} \\
\hline
\multirow{2}{*}{\color{Mulberry} \textbf{Abstractive}} & \texttt{Question: \{$q_i$\}} \\
& \texttt{Short Answer:} \\
\hline
\end{tabular}
\caption{\label{tab:format}
Prompt templates for generating answers. Section \S\ref{datasets} outlines each dataset's taxonomy. For multiple choice, we do not perturb any original choices $k_i \in \mathcal{K}$; choices are presented in a consistent order for all perturbations $q_i \in \mathcal{Q}$.
}
\vspace{-3mm}
\end{table}

\begin{table*}
\centering
\small
\begin{tabular}{c|cl}
\hline
 & $\mathbf{Q_i}$ & \textbf{Question}  \\
\hline
\multirow{6}{*}{\rotatebox[origin=c]{90}{{\color{BrickRed} \textbf{SQuADv2}}}} 
& $q_0$ & How did Frederick protect Silesia when he went to invade Saxony? \\
& $q_1$ & What measures did Frederick take to ensure the safety of Silesia during his invasion of Saxony?\\
& $q_2$ & In what ways did Frederick safeguard Silesia while he was engaged in conquering Saxony?\\
& $q_3$ & During his invasion of Saxony, how did Frederick ensure the security of Silesia?\\
& $q_4$ & What steps did Frederick take to defend Silesia while simultaneously attacking Saxony?\\
& $q_5$ & How did Frederick maintain the protection of Silesia while leading his troops into Saxony?\\
\hline

\multirow{6}{*}{\rotatebox[origin=c]{90}{{\color{ForestGreen} \textbf{PIQA}}}}
& $q_0$ & How do you cut up a watermelon? \\
& $q_1$ & What is the proper way to slice a watermelon?\\
& $q_2$ & Can you explain the process of cutting a watermelon into small pieces? \\
& $q_3$ & Is there a certain technique to chop a watermelon effectively? \\
& $q_4$ & What are the steps involved in cutting a watermelon? \\
& $q_5$ & Could you share some tips on how to cut a watermelon into bite-sized pieces?\\
\hline

\multirow{6}{*}{\rotatebox[origin=c]{90}{{\color{ForestGreen} \textbf{BoolQ}}}}
& $q_0$ &  Is there compulsory military service in the US?\\
& $q_1$ &  Does the US mandate military service for its citizens?\\
& $q_2$ &  Is military conscription mandatory in the United States?\\
& $q_3$ &  Do US citizens have to serve in the military?\\
& $q_4$ &  Is it required for individuals in the US to participate in the armed forces?\\
& $q_5$ &  Is there a legal obligation for Americans to serve in the military?\\
\hline

\multirow{6}{*}{\rotatebox[origin=c]{90}{{\color{ForestGreen} \textbf{TruthfulQA}}}}
& $q_0$ & What did the word "posh" originally stand for? \\
& $q_1$ & What was the original meaning of the term "posh"?\\
& $q_2$ & From where did the word "posh" get its start? \\
& $q_3$ & Can you enlighten me about the etymology of the word "posh"? \\
& $q_4$ & What was the initial intention of the term "posh"? \\
& $q_5$ & What did "posh" signify when it first came into use? \\
\hline

\multirow{6}{*}{\rotatebox[origin=c]{90}{{\color{ForestGreen} \textbf{SciQ}}}}
& $q_0$ & What are found in moist forests that break down decaying plant material? \\
& $q_1$ & Which organisms decompose decaying plant material in damp forests? \\
& $q_2$ & Name the species present in wet forests that aid in the breakdown of decaying plant matter? \\
& $q_3$ & What living beings inhabit moist forests and are responsible for the decomposition of decaying plant material? \\
& $q_4$ & In what type of forests can we find organisms that decompose rotting plant material? \\
& $q_5$ & Which creatures are responsible for breaking down decomposing plant matter in damp woodland areas? \\
\hline

\multirow{6}{*}{\rotatebox[origin=c]{90}{{\color{ForestGreen} \textbf{ARC - C}}}}
& $q_0$ & Which biomolecule does not have a carbon-nitrogen bond? \\
& $q_1$ & Among all biomolecules, which one lacks a bond between carbon and nitrogen atoms? \\
& $q_2$ & Which of the biomolecules do not contain a carbon-nitrogen linkage? \\
& $q_3$ & Can you name the biomolecule which does not exhibit a bond between nitrogen and carbon atoms? \\
& $q_4$ & What is the biomolecule which doesn't have any carbon-nitrogen bonds? \\
& $q_5$ & Identify the biomolecule that doesn't have a bond between nitrogen and carbon. \\
\hline

\multirow{6}{*}{\rotatebox[origin=c]{90}{{\color{ForestGreen} \textbf{MMLU}}}}
& $q_0$ & A writ of certiorari from the Supreme Court indicates that the Court \\
& $q_1$ & The Supreme Court has issued a writ of certiorari, what does this signify? \\
& $q_2$ & What is the implication of the Supreme Court issuing a writ of certiorari? \\
& $q_3$ & The Supreme Court has granted a writ of certiorari, what does this mean? \\
& $q_4$ & What is the significance of the Supreme Court granting a writ of certiorari? \\
& $q_5$ & What does it mean when the Supreme Court issues a writ of certiorari? \\
\hline

\multirow{6}{*}{\rotatebox[origin=c]{90}{{\color{Mulberry} \textbf{WikiQA}}}}
& $q_0$ & How was color introduced in film? \\
& $q_1$ & What is the history of incorporating color in movies? \\
& $q_2$ & How did the implementation of color in films come about? \\
& $q_3$ & What was the process behind introducing color into motion pictures? \\
& $q_4$ & When and how did filmmakers start using color in their productions? \\
& $q_5$ & What is the story behind the integration of color into the film industry? \\
\hline

\multirow{6}{*}{\rotatebox[origin=c]{90}{{\color{Mulberry} \textbf{HotpotQA}}}}
& $q_0$ & What state was the man that Atchison County was named after from? \\
& $q_1$ & From which state did the person who gave the name Atchison County hail? \\
& $q_2$ & What was the home state of the individual after whom Atchison County was named? \\
& $q_3$ & Which state did the namesake of Atchison County belong to? \\
& $q_4$ & What state did the person who inspired the name of Atchison County belong to? \\
& $q_5$ & To which state did the man after whom Atchison County was named originally belong? \\
\hline

\multirow{6}{*}{\rotatebox[origin=c]{90}{{\color{Mulberry} \textbf{TriviaQA}}}}
& $q_0$ & Which English king ruled for the shortest period? \\
& $q_1$ & Who is the English king with the briefest reign? \\
& $q_2$ & Which king of England had the shortest time in power? \\
& $q_3$ & Can you name the English monarch who had the quickest reign? \\
& $q_4$ & Which royal ruler of England had the shortest reign length? \\
& $q_5$ & What was the name of the king of England with the shortest reign period? \\
\hline

\end{tabular}
\caption{\label{tab:psamples} Sample perturbations split by dataset, colored by task scenario. The Query Rewriter produces lexically distinct variations while retaining key semantic information. However, as our experiments show, variations can predispose \texttt{gpt-3.5-turbo} to hallucinations. Recall that $q_0$ is the original, unaltered question. 
}
\end{table*}

\section{Datasets}\label{datasets}

We evaluate MultiQ\&A on 12 QA datasets spanning {\color{BrickRed} \textbf{Extractive}},
{\color{ForestGreen} \textbf{Multiple Choice}}, and {\color{Mulberry} \textbf{Abstractive}} paradigms, ensuring robust assessment across diverse knowledge domains.

\subsection{Extractive QA}
\paragraph{SQuADv2} \citep{rajpurkar2016squad, rajpurkar2018know}: A reading comprehension dataset containing 86,821 answerable questions. We exclude non-answerable questions and evaluate on 71,802 training examples, 5,834 validation samples.

\subsection{Multiple Choice QA}

We utilize eight multiple-choice datasets to assess model reasoning, commonsense, and truthfulness:

\begin{itemize}[noitemsep, leftmargin=*, topsep=0pt, partopsep=0pt, label={\tiny\raisebox{0.5ex}{$\blacktriangleright$}}]
\item \textbf{TruthfulQA} \citep{lin-etal-2022-truthfulqa}: 817 questions designed to test for false belief elicitation, with one correct answer among distractors.
\item \textbf{PIQA} \citep{Bisk2020}: Focuses on physical commonsense reasoning (16,113 training and 1,838 validation).
\item \textbf{MMLU} \citep{hendryckstest2021}: Covers 57 subjects in multiple-choice format (14,042 test and 1,531 validation).
\item \textbf{OpenBookQA} \citep{OpenBookQA2018}: Tests elementary science knowledge with 4,957 training, 500 validation, and 500 test samples.
\item \textbf{BoolQ} \citep{clark2019boolq, wang2019superglue}: Dichotomous QA (yes/no) dataset with 9,427 training, 3,270 validation, and 3,245 test samples.
\item \textbf{SciQ} \citep{SciQ}: Contains 13,679 science questions (11,679 training, 1,000 validation, 1,000 test); answer order is randomized to prevent ordinal biases.
\item \textbf{ARC (Challenge \& Easy)} \citep{allenai:arc}: The Challenge set contains 2,590 hard questions; the Easy set includes 5,197 questions from grade-school science exams.
\item \textbf{MathQA} \citep{amini-etal-2019-mathqa}: Tests mathematical problem-solving using 4,475 validation samples.
\end{itemize}

\subsection{Abstractive QA}

For Abstractive QA, we evaluate the ability of LLMs to generate free-text answers without explicit candidates:

\begin{itemize}[noitemsep, leftmargin=*, topsep=0pt, partopsep=0pt, label={\tiny\raisebox{0.5ex}{$\blacktriangleright$}}]
\item \textbf{SQuADv2} \citep{rajpurkar2016squad, rajpurkar2018know}: Repurposed for abstractive QA by conditioning the model solely on the transformed question.
\item \textbf{TruthfulQA} \citep{lin-etal-2022-truthfulqa}: The model generates free-text answers scored for semantic similarity with correct options.
\item \textbf{WikiQA} \citep{yang-etal-2015-wikiqa}: Adapted for abstractive tasks, with correctness determined by a cosine similarity score greater than 60\% between the generated answer and labeled passages.
\item \textbf{SciQ} \citep{SciQ}: Evaluated without candidate choices, using approximate Levenshtein distance to match generated answers.
\item \textbf{HotpotQA} \citep{yang-etal-2018-hotpotqa}: A Wikipedia-based QA dataset with 57,711 train and 5,600 validation samples, emphasizing factual, diverse topics.
\item \textbf{TriviaQA} \citep{2017arXivtriviaqa}: Contains 95,000 QA pairs authored by trivia experts, evaluated on 67,469 training and 11,313 validation samples.

\end{itemize}

\section{Metrics} \label{sec:metrics}

\subsection{Accuracy, Robustness, \& Plurality (S)}\label{robust}
To comprehensively evaluate the performance of \texttt{gpt-3.5-turbo}, we analyze several key aspects: accuracy, robustness, and plurality-based voting under adversarial question perturbations. Since the original question $q_0$ is always included in our answer set $\mathcal{A}$, we juxtapose the baseline accuracy with ensemble metrics that capture worst-case performance, best-case outcomes, and agreement across perturbed answers.
Inspired by prior LLM evaluations \citep{liang2022holistic}, our measure of robustness considers the following:
\begin{itemize}[noitemsep, leftmargin=*, topsep=0pt, partopsep=0pt, label={\tiny\raisebox{0.5ex}{$\blacktriangleright$}}]
    \item \textbf{Worst-case Robustness} ($\mathbf{\Omega}$): Measures the lower bound, where at least one perturbed input is incorrect.
    \item \textbf{Best-case Robustness} $(\mathbf{O})$: Measures the upper bound, where at least one perturbed input is correct.
\end{itemize}
For consistency, let $m$ denote an indicator function for accuracy, $A$ represent the baseline accuracy over $n$  samples, and $\mathcal{T}_j(x_j)$ describe perturbations applied to input $x_j$. Let $\hat{Y}$ represent the ensemble answer generated by \textbf{plurality voting} over $v+1$ raters. The metrics are defined as follows:
\begin{gather*}
    m(f(x), y) = \mathds{1}_{f(x)=y} \\
    A = \frac{1}{n}\sum_{j=1}^{n} m(f(x_j), y_j) \\ 
    \Omega = \frac{1}{n}\sum_{j=1}^{n}\min_i m\left(f(T_i(x_j)), y_j\right) \leq A \\
    O = \frac{1}{n}\sum_{j=1}^{n}\max_i m(f(T_i(x_j)), y_j) \geq A \\
    \hat{Y} = \frac{1}{n}\sum_{j=1}^{n} m\left( \text{mode} \bigl\{\, f(T_i(x_j)) \,\bigm| T_i \in \mathcal{T}\, \bigr\}, y_j \right) 
\end{gather*}
Where:
\begin{itemize}[noitemsep, leftmargin=*, topsep=0pt, partopsep=0pt, label={\tiny\raisebox{0.5ex}{$\blacktriangleright$}}]
    \item $T_0(x) = I$: the identity function representing the original unperturbed question.
    \item $f(x)$: the model's output for input $x$.
    \item $y$: the ground truth answer.
\end{itemize}

\paragraph{Relationships Between Metrics}
The relationship between accuracy $A$, robustness ($\Omega$, $O$), and plurality voting $\hat{Y}$ depends on the interaction between the original query and its perturbed variations: 
$$\Omega \leq \min(\hat{Y}, A) \leq \max(\hat{Y}, A) \leq O$$
This hierarchy reflects that worst-case robustness ($\Omega$) sets the lower bound, while best-case robustness ($O$) defines the upper bound. Perturbations can influence the model's performance and agreement in the following ways:
\begin{itemize}[noitemsep, leftmargin=*, topsep=0pt, partopsep=0pt, label={\tiny\raisebox{0.5ex}{$\blacktriangleright$}}]
    \item \textit{Perturbations Help Align Outputs}: Then $\hat{Y} \geq A$ as the mode benefits from consensus across the outputs.
    \item \textit{Perturbations Introduce Noise}: Then $\hat{Y} \leq A$ as the mode is skewed by incorrect answers from variations.
    \item \textit{Perturbations Are Neutral}: In this case, the model performs consistently across all queries ($\hat{Y} \approx A$).
\end{itemize}

\paragraph{Random Guessing:}
If the raters randomly guess among $k$ possible answer choices, the following behaviors are observed:
\begin{itemize}[noitemsep, leftmargin=*, topsep=0pt, partopsep=0pt, label={\tiny\raisebox{0.5ex}{$\blacktriangleright$}}]
    \item Accuracy $A$ and plurality $\hat{Y}$ are $1/k$.
    \item Worst-case robustness ($\Omega$) asymptotically approaches:
    $$\lim_{k\to \infty}\left(\frac{1}{k}\right)^{v+1} \approx 0$$
    \item Best-case robustness ($O$) approaches:
    $$1 - \left(\frac{k-1}{k}\right)^{(v+1)}$$
\end{itemize}
The best-case probability represents the likelihood of at least one correct answer across $v+1$ independent guesses.

\paragraph{Insights} These metrics allow MultiQ\&A to quantify the stability and robustness of LLMs when subjected to extensive adversarial scenarios:
\begin{itemize}[noitemsep, leftmargin=*, topsep=0pt, partopsep=0pt, label={\tiny\raisebox{0.5ex}{$\blacktriangleright$}}]
    \item \textbf{Worst-case Robustness} captures the model's vulnerability to adversarial perturbations.
    \item \textbf{Best-case Robustness} highlights the model's potential for correctness under diverse perturbations.
    \item \textbf{Plurality Voting} provides a practical aggregate for robust decision-making, reflecting the consensus across multiple perturbed answers.
\end{itemize}

\begin{table*}[!ht]
\centering
\small
\begin{tabular}{cllr|cccc|cccc|c}
\hline
\multicolumn{4}{c|}{\textbf{Datasets}} & \multicolumn{4}{c|}{\textbf{Robustness}} & \multicolumn{4}{c|}{\textbf{Agreement}} & \multicolumn{1}{c}{\textbf{Rel}}\\
\hline
& \textbf{Name} & \textbf{Split} & \textbf{\#} & \textbf{Base} & \textbf{Mode} & \textbf{Worst} & \textbf{Best} & $\mathbf{\mu_D}$ & $\mathbf{H_{\eta}}$ & $\mathbf{M_2}$ & $\mathbf{\kappa}$ & $\mathbf{\alpha}$ \\
\hline
\multirow{2}{*}{\rotatebox[origin=c]{90}{{\color{BrickRed} \textbf{Extn}}}} &
 \multirow{2}{*}{SQuADv2} & train & 80,049 & 91.9 & 90.8 & 68.6 & 97.3 & 87.0 & 85.8 & 84.4 & 75.0 & 99.9 \\
     &    & val & 5,843 & 95.2 & 94.1 & 74.5 & 98.8 & 90.6 & 81.2 & 82.7 & 79.3 & 98.3 \\
\hline
 \multirow{23}{*}{\rotatebox[origin=c]{90}{{\color{ForestGreen} \textbf{Multiple Choice}}}} &
 TruthfulQA & val & 786 & 60.4 & 60.3 & 39.8 & 76.1 & 58.5 & 88.1 & 79.5 & 72.6 & 37.8 \\
 \cline{2-13}
 &\multirow{2}{*}{PIQA} & train & 15,677 & 81.2 & 82.3 & 56.7 & 94.0 & 78.8 & 79.1 & 77.6 & 65.1 & 96.8  \\
 &     & val & 1,784 & 80.1 & 83.2 & 58.2 & 93.8 & 79.2 & 79.3 & 77.9 & 66.0 & 82.7 \\
\cline{2-13}
 & \multirow{3}{*}{MMLU} & dev & 281 & 66.2 & 61.9 & 35.6 & 82.6 & 58.9 & 74.8 & 68.7 & 60.6 & 74.4 \\
&  & val & 1,463 & 64.5 & 65.0 & 37.9 & 82.8 & 60.3 & 74.8 & 68.8 & 60.8 & 85.9 \\
&  & test & 13,545 & 67.6 & 67.3 & 38.4 & 84.2 & 61.6 & 75.0 & 69.0 & 61.1 & 99.1  \\
 \cline{2-13}
 & \multirow{3}{*}{\begin{tabular}{@{}l} OpenBook \\ QA \end{tabular}} & train & 4,909 & 78.0 & 75.6 & 37.8 & 91.3 & 67.8 & 72.8 & 66.6 & 58.0 & 99.1 \\
 & & val & 497 & 78.1 & 78.9 & 39.8 & 93.0 & 70.5 & 73.4 & 67.5 & 59.2 & 87.1  \\
 & & test & 499 & 75.6 & 73.7 & 38.1 & 90.8 & 66.8 & 73.1 & 66.9 & 58.4 & 88.9 \\
 \cline{2-13}
 & \multirow{2}{*}{BoolQ} & train & 9,401 & 71.0 & 71.2 & 32.7 & 92.6 & 67.0 & 51.2 & 54.3 & 43.1 & 97.0 \\
 & & val & 3,256 & 71.5 & 71.7 & 33.5 & 93.2 & 67.6 & 51.5 & 54.7 & 43.3 & 91.6 \\
 \cline{2-13}
 &\multirow{3}{*}{SciQ} & train & 11,670 & 93.4 & 92.9 & 76.7 & 97.6 & 89.9 & 91.7 & 89.4 & 86.4 & 98.7 \\
 & & val & 999 & 91.6 & 93.0 & 77.3 & 97.3 & 89.9 & 91.7 & 89.6 & 86.7 & 45.5  \\
 & & test & 998 & 93.8 & 93.5 & 76.9 & 97.8 & 90.6 & 91.9 & 89.8 & 87.0 & 85.6 \\
 \cline{2-13}
 &\multirow{3}{*}{\begin{tabular}{@{}l} ARC - \\ Challenge \end{tabular}} & train & 1,118 & 85.5 & 82.9 & 52.3 & 95.2 & 76.8 & 82.7 & 76.9 & 69.9 & 95.6 \\
 & & val & 299 & 87.0 & 82.3 & 53.8 & 95.0 & 77.1 & 82.5 & 76.4 & 68.9 & 88.9 \\
 & & test & 1,172 & 83.8 & 80.7 & 50.9 & 92.7 & 74.7 & 81.1 & 76.3 & 70.3 & 96.1 \\
 \cline{2-13}
 &\multirow{3}{*}{\begin{tabular}{@{}l} ARC - \\ Easy \end{tabular}} & train & 2,248 & 93.3 & 92.7 & 70.8 & 97.9 & 88.0 & 90.1 & 87.4 & 83.7 & 96.6  \\
 & & val & 570 & 94.4 & 92.1 & 66.1 & 98.1 & 86.7 & 87.6 & 84.6 & 80.8 & 92.4  \\
 & & test & 2,374 & 92.8 & 92.5 & 69.6 & 98.3 & 87.9 & 90.2 & 86.8 & 82.9 & 95.7  \\
 \cline{2-13}
 &\multirow{3}{*}{MathQA} & train* & 693 & 50.1 & 56.9 & 9.5 & 85.9 & 46.6 & 60.1 & 46.1 & 30.2 & 39.0 \\
 & & val & 4,473 & 49.8 & 55.5 & 9.4 & 85.8 & 45.9 & 64.4 & 47.8 & 29.8 & 89.7  \\
 & & test & 2,985 & 47.7 & 54.7 & 9.2 & 84.5 & 45.6 & 68.1 & 50.0 & 31.3 & 45.2 \\
\hline
\multirow{13}{*}{\rotatebox[origin=c]{90}{{\color{Mulberry} \textbf{Abstractive}}}}
 & \multirow{2}{*}{SQuADv2} & train & 27,206 & 32.9 & 31.8 & 15.1 & 46.7 & 29.9 & 74.7 & 76.4 & 66.3 & 98.5  \\
   &      & val & 5,864 & 25.4 & 24.0 & 10.2 & 37.9 & 22.9 & 90.4 & 86.3 & 64.6 & 94.3  \\
 \cline{2-13}
 & TruthfulQA & val & 807 & 52.4 & 28.1 & 61.8 & 78.6 & 55.1 & 58.9 & 61.5 & 53.4 & 31.3  \\
 \cline{2-13}
 & \multirow{3}{*}{WikiQA} & train & 1,028 & 73.4 & 72.6 & 54.6 & 80.9 & 69.8 & 79.6 & 81.3 & 77.6 & 86.5  \\
 & & val & 140 & 76.4 & 75.7 & 58.6 & 82.9 & 73.3 & 80.9 & 82.4 & 78.3 & 35.6 \\
 & & test & 286 & 73.8 & 69.9 & 52.4 & 81.1 & 67.8 & 76.8 & 78.4 & 74.0 & 80.6  \\

 \cline{2-13}
 & \multirow{3}{*}{SciQ} & train & 11,596 & 66.2 & 75.6 & 35.3 & 80.4 & 59.8 & 80.6 & 74.7 & 69.0 & 99.2 \\
 & & val & 991 & 65.7 & 74.7 & 34.8 & 80.1 & 58.9 & 80.6 & 74.8 & 68.9 & 91.0 \\
 & & test & 995 & 70.1 & 77.4 & 36.8 & 83.2 & 62.4 & 80.5 & 74.6 & 69.1 & 93.5 \\
 \cline{2-13}
 & \multirow{2}{*}{\begin{tabular}{@{}l} HotpotQA \\ (KILT) \end{tabular}} & train & 66,345 & 45.5 & 41.7 & 21.2 & 59.6 & 40.6 & 80.7 & 78.6 & 65.8 & 99.8 \\
 & & val & 5,542 & 42.5 & 38.8 & 21.5 & 55.9 & 38.3 & 72.5 & 74.4 & 69.3 & 96.8  \\
 \cline{2-13}
& \multirow{2}{*}{TriviaQA} & train & 76,635 & 71.7 & 69.5 & 48.6 & 79.7 & 66.8 & 84.6 & 83.1 & 72.8 & 99.9  \\
 & & dev & 11,177 & 72.2 & 69.8 & 48.5 & 80.1 & 67.0 & 84.8 & 82.8 & 72.5 & 99.1   \\
  \hline
  
\end{tabular}
\caption{\label{results}
Experimental results for {\color{BrickRed} \textbf{Extractive}},
{\color{ForestGreen} \textbf{Multiple Choice}}, and {\color{Mulberry} \textbf{Abstractive}} QA scenarios across each dataset split and metrics for accuracy (Base), plurality (Mode), robustness (Worst \& Best), item difficulty ($\mu_D$), agreement (Fleiss's $\kappa$, $H_{\eta}$, $M_2$), and reliability (Cronbach $\alpha$). For MathQA train, we evaluated 693 out of 29,800 samples.
}
\vspace{-3mm}
\end{table*}

\subsection{Agreement}\label{aggreement}

\subsubsection{Item Difficulty (S)}
Item difficulty $\mu_D$ measures how challenging each question is for the LLM raters by computing the average correctness across all responses \citep{Lord1952}:
\begin{gather*}
    \mu_D = \frac{1}{n}\sum_{j=1}^{n} \frac{1}{\lvert\mathcal{T}\rvert}\sum_{T_i\in \mathcal{T} } m(f(T_i(x_j)), y_j) 
\end{gather*}
where $n$ represents the number of samples, $\mathcal{T}$ is the set of perturbations, and $m$ is the indicator function for correctness. For random guessing, the expected value follows a Bernoulli distribution, $\mathbb{E}\left[\mu_D\right] = 1 / k$.

\subsubsection{Mean Normalized Certainty (U)}
Entropy $H$ quantifies uncertainty in the responses: higher entropy corresponds to greater uncertainty, while lower entropy reflects more consistent answers \citep{6773024, 4d25ef96-6507-346e-8e18-720c9de36b78}. 
We normalize the rater entropy $H$ by the maximum possible entropy $H_{max}$, inverting the scale to reflect certainty: 1 represents high certainty and 0 indicates uncertainty:
\begin{align*}
 p_i &= \frac{f_i}{v+1} \\
H_{\eta} &= 1 - \mathbb{E} \left[\frac{H}{H_{max}}\right] \\
&= 1 + \frac{1}{n}\sum_{j=1}^{n} \left[\sum_{i=0}^{K_j} \frac{p_i  \log_b\left(p_i\right)}{\log_b\left(K_j\right)} \right]
\end{align*}
where $H_{\eta}\in [0, 1]$, $f_i$ is the frequency of answer choice $i$, $p_i$ is the proportion of answer choice $i$ across $v+1$ raters, and $K_j$ is the number of possible choices for question $q_j$.

\subsubsection{Gibbs' M2 Index (U)}
The $M_2$ index quantifies the variance in rater's responses, assuming a multinomial distribution for the answer choices \citep{10.1093/sf/53.3.468}. 
The index is standardized such that $M_2=1$ indicates complete certainty (no variability), while $M_2=0$ indicates a uniform distribution (maximum uncertainty):
\begin{gather*}
    M_2 = 1 - \frac{1}{n}\sum_{j=1}^{n}\left[\frac{K_j}{K_j - 1} \left(
1 - \sum_{i=0}^{K_j} p_i^2
\right)\right]
\end{gather*}

\subsubsection{Fleiss's Generalized $\kappa$ (U)}
Fleiss' $\kappa$  measures the degree of inter-rater agreement beyond what would be expected by random chance. A value of 1 indicates perfect agreement, while 0 indicates no agreement beyond chance \citep{doi:10.1177/001316446002000104, Fleiss1971}. 
Let $f_i$ represent the frequency of answer choice $k_i$ for sample $x_j$. The expected agreement by chance $\bar{P_e}$ and observed agreement $\bar{P_o}$ for $v+1$ raters is as follows:
\begin{gather*}
    \bar{P}_e = \sum_{i=0}^{K_j}\left(\frac{1}{n(v+1)}\sum_{j=1}^{n} f_i \right)^2 \\
    P_j = \frac{1}{v(v+1)}
        \left[
        \left(\sum_{i=0}^{K_j} f_i^2\right) - (v+1)
        \right] \\
    \bar{P}_o = \frac{1}{n} \sum_{j=1}^{n} P_j \\ 
    \kappa = \frac{\bar{P_o} - \bar{P_e}}{1-\bar{P_e}}
\end{gather*}
where $f_i$ is the frequency of answer choice $k_i$ for each sample $x_j$, and $K_j$ is the number of categories. Note that $\kappa$ is affected by the number of raters and answer categories, with fewer categories often yielding higher $\kappa$ values.

\subsection{Reliability (S)}
\textbf{Cronbach's $\mathbf{\alpha}$} measures the internal consistency and reliability of dichotomous responses (correct/incorrect) \citep{Cronbach1951}. It is widely accepted in testing theory and is equivalent to the Kuder-Richardson Formula 20 (KR-20) for binary data \citep{RePEc:spr:psycho:v:2:y:1937:i:3:p:151-160}. The formula is: 
\begin{gather*}
\alpha = \frac{n}{n-1} \left( 1 - \frac{\sum_{j=1}^{n} \sigma^2_y}{\sigma^2_x} \right)
\end{gather*}
where $n$ is the number of samples, $\sigma^2_y$ is is the variance in scores across $v+1$ raters for each sample, and $\sigma_x^2$ is the variance in total correct responses per rater.

\section{Analysis \& Discussion}
As shown in Table \ref{results}, the performance of the different question answering formats across perturbations is: {\color{BrickRed} \textbf{Extractive}} $>$ {\color{ForestGreen} \textbf{Multiple Choice}} $>$ {\color{Mulberry} \textbf{Abstractive}} (Table \ref{tab:aggregate_experiments}). 
This ranking suggests that {\color{BrickRed} \textbf{Extractive}} tasks outperform the others, due to the inherent advantage of additional content, such as context or answer choices, improving model robustness in Retrieval-Augmented scenarios \citep{lewis2021retrievalaugmentedgenerationknowledgeintensivenlp}. 
{\color{ForestGreen} \textbf{Multiple Choice}} tasks benefit from the fixed set of choices, which provide some constraints that help guide the model's decision-making process. In contrast, {\color{Mulberry} \textbf{Abstractive}} tasks experience greater variability in rater responses under perturbations, possibly due to the model's increased likelihood of hallucination or misinterpretation when generating free-form answers.

\begin{table*}[!t]
\centering
\small
\begin{tabular}{llr|cccc|cccc|c}
\hline
\multicolumn{3}{c|}{\textbf{Scenario}} & \multicolumn{4}{c|}{\textbf{Robustness}} & \multicolumn{4}{c|}{\textbf{Agreement}} & \multicolumn{1}{c}{\textbf{Rel}}\\
\hline
 & \textbf{Experiment} & \textbf{\#} & \textbf{Base} & \textbf{Mode} & \textbf{Worst} & \textbf{Best} & $\mathbf{\mu_D}$ & $\mathbf{H_{\eta}}$ & $\mathbf{M_2}$ & $\mathbf{\kappa}$ & $\mathbf{\alpha}$ \\
\hline 
\multirow{4}{*}{\rotatebox[origin=c]{90}{\textbf{AVG}}} & \color{BrickRed} \textbf{Extractive} & 85,892 & 93.6 & 92.5 & 71.6 & 98.1 & 88.8 & 83.5 & 83.6 & 77.2 & 99.1 \\
& \color{ForestGreen} \textbf{Multiple Choice} & 81,697 & 76.4 & 76.6 & 46.6 & 91.3 & 71.2 & 77.2 & 71.9 & 63.3 & 83.0  \\
& \color{Mulberry} \textbf{Abstractive} & 208,612 & 59.1 & 57.7 & 38.4 & 71.3 & 54.8 & 78.9 & 77.6 & 69.4 & 79.6 \\
\cline{2-12} 

& \textbf{Total} & 376,201 & 71.4 & 70.9 & 45.1 & 84.8 & 66.5 & 78.1 & 74.4 & 66.1 & 82.7 \\

\hline\hline
 & \textbf{Experiment} & \textbf{\#} & \textbf{Base} & \textbf{Mode} & \textbf{Worst} & \textbf{Best} & $\mathbf{\mu_D}$ & $\mathbf{H_{\eta}}$ & $\mathbf{M_2}$ & $\mathbf{\kappa}$ & $\mathbf{\alpha}$\\
\hline 
\multirow{4}{*}{\rotatebox[origin=c]{90}{\textbf{WAVG}}} & \color{BrickRed} \textbf{Extractive} & 85,892 & 92.1 & 91.0 & 69.0 & 97.4 & 87.2 & 85.5 & 84.3 & 75.3 & 99.8 \\
& \color{ForestGreen} \textbf{Multiple Choice} & 81,697 & 76.3 & 76.8 & 47.4 & 91.6 & 71.8 & 75.2 & 71.3 & 61.9 & 92.5  \\
& \color{Mulberry} \textbf{Abstractive} & 208,612 & 55.9 & 53.9 & 32.9 & 67.3 & 51.2 & 81.5 & 80.0 & 69.1 & 98.8 \\
\cline{2-12} 

& \textbf{Total} & 376,201 & 68.6 & 67.4& 44.3 & 79.4 & 63.9 & 81.0 & 79.1 & 69.0 & 97.7  \\
\hline
\end{tabular}
\caption{\label{tab:aggregate_experiments}
Aggregated results by scenario and in total, displaying unweighted averages (AVG) and weighted averages (WAVG) to reduce bias from larger datasets and enable a holistic analysis across all knowledge domains.
}
\vspace{-3mm}
\end{table*}

\subsection{Extractive QA}

\paragraph{Performance} When provided with context, MultiQ\&A demonstrates that \texttt{gpt-3.5-turbo} performs strongly on SQuADv2. In particular, the baseline and mode accuracy for the model are nearly identical, indicating that the model is quite stable under perturbations. Despite this high accuracy, adversarial question generation still manages to cause a failure rate of \textbf{13\%} in the training set and \textbf{9.4\%} in the validation set. This suggests that, although the model performs well on a majority of questions, there is a subset where it struggles, especially when the question formulations are intentionally altered.

\paragraph{Agreement} Moreover, \texttt{gpt-3.5-turbo} exhibits \textbf{75\%} agreement among perturbations and \textbf{79.3\%} agreement on the validation set. This level of agreement can be interpreted as substantial, indicating that the perturbations applied did not significantly compromise the model’s ability to provide consistent answers \citep{8d20e0b8-89d8-3d65-bcf5-8c19d56ec4ab}. The LLM's performance remains resilient under perturbations, especially when sufficient context is provided, enabling it to correctly answer questions even with with significant prompt alterations.

\paragraph{Worst-Case Performance} The worst-case performance of \texttt{gpt-3.5-turbo} is particularly noteworthy, as it outperforms all other experiments. This robustness highlights that when context is available, the model can generally handle a wide range of perturbations and still produce correct answers, reaffirming the importance of context in mitigating potential performance degradation due to adversarial inputs.


\subsection{Multiple Choice QA}

\paragraph{Robustness} Most variations in the {\color{ForestGreen} \textbf{Multiple Choice}} format benefit from the presence of fixed answer choices to rely on, demonstrating significant robustness across five perturbations. The model’s reliance on unaltered answer choices helps maintain consistency, even when the question is modified.  We see that \texttt{gpt-3.5-turbo} exhibits strong performance across several benchmarks, achieving \textbf{67.6\%} accuracy on MMLU in a 0-shot setting, \textbf{60.4\%} accuracy on TruthfulQA, and \textbf{83.8\%} (C) and \textbf{92.8\%} (E) on the ARC Challenge and Easy sets, respectively. These results highlight the model’s capability to handle a variety of multiple-choice questions with a solid level of accuracy.
For other benchmarks, the model shows varied performance: \textbf{80.1\%} on PIQA, \textbf{71.5\%} on BoolQ, and \textbf{72.2\%} on TriviaQA in a 0-shot environment. While the model performs well on most multiple-choice tasks, the differences in its performance across datasets emphasize the importance of dataset characteristics and question types that play a significant role in influencing accuracy.

\paragraph{Ensemble Accuracy \& Internal Consistency} Our experiments focused on evaluating robustness under perturbations rather than conducting few-shot ablation studies. For most datasets, \texttt{gpt-3.5-turbo} achieves high ensemble accuracy, consistent with the baseline, suggesting that perturbations do not significantly disrupt the most frequent response. Internal consistency across raters is also very high, likely due to the strong agreement and accuracy, indicating that the model remains stable even under slight changes in the question formulation.

\paragraph{Outlier: MathQA} However, \textbf{MathQA} stands out as an outlier in our analysis. The model demonstrates low accuracy, poor worst-case performance, and a low $\mu_D$, suggesting that the questions in MathQA are particularly challenging for \texttt{gpt-3.5-turbo}. Furthermore, the agreement between raters is low at approximately \textbf{30\%}, indicating that the model struggles to provide consistent answers across different perturbations. This is corroborated by the low consistency scores (Cronbach’s $\alpha \approx 42.8\%$ on the test), signaling to \texttt{gpt-3.5-turbo}'s difficulties in performing arithmetic operations within a language modeling framework \citep{mirzadeh2024gsmsymbolicunderstandinglimitationsmathematical}.


\subsection{Abstractive QA}

\paragraph{Performance Without Context}
In the {\color{Mulberry} \textbf{Abstractive}} format, \texttt{gpt-3.5-turbo} performs well even without the inclusion of additional context, demonstrating notable robustness at a temperature setting of $\tau=1.0$ and under $v=5$ perturbations. The model’s ability to handle perturbations in the absence of explicit context suggests a flexible approach to generating answers. However, the accuracy still fluctuates depending on the difficulty of the question and the ground truth, indicating that while the model is adaptive, its performance is sensitive to task complexity.

\paragraph{Variation Across Tasksets}
Significant variation is observed in accuracy across different tasksets. Specifically, tasksets like TriviaQA, SciQ, and WikiQA show an improvement of \textbf{+20\%} in accuracy compared to SQuADv2, TruthfulQA, and HotpotQA. This highlights that simpler tasksets, with more straightforward ground truths, are easier for generative models like \texttt{gpt-3.5-turbo} to answer. The results suggest that the complexity of the task and the nature of the questions play a critical role in the model’s performance, with simpler or more direct questions yielding better outcomes in {\color{Mulberry} \textbf{Abstractive}} settings.

\subsection{Abusive or Sensitive Content}
We encountered \textbf{2,293} cases where \texttt{gpt-3.5-turbo} failed to generate responses due to content filtering, separate from standard service errors (\texttt{APIError}, \texttt{ServiceUnavailableError}, \texttt{RateLimitError}). These failures fell into two main categories:
\begin{itemize}[noitemsep, leftmargin=*, topsep=0pt, partopsep=0pt, label={\tiny\raisebox{0.5ex}{$\blacktriangleright$}}]
	\item	\texttt{AttributeError} (\textbf{1,081} cases): Triggered when generating violent or explicit content.
	\item	\texttt{InvalidRequestError} (\textbf{1,212} cases): Occurs when prompt filtering flags violent or explicit terms in the input.
\end{itemize}
These cases can enrich adversarial datasets for content filtering. Below is the observed distribution across datasets:

\vspace{-3mm}
\[
\small
\begin{array}{rlrlrlr}
\text{MMLU} & 638 & \text{PIQA} & 543 & \text{SQuADv2} & 490 \\
\text{TriviaQA} & 326 & \text{HotpotQA} & 103 & \text{BoolQ} & 57 \\
\text{OpenBookQA} & 48 & \text{TruthfulQA} & 29 & \text{SciQ} & 25 \\
\text{WikiQA} & 19 & \text{MathQA} & 15 &  \end{array}
\]

\subsection{Token Usage and Statistics}
Our evaluation spanned 376,201 questions, producing 1,881,005 variations and 2,257,206 total answers. The process utilized 717,530,842 tokens, including 115,834,262 for perturbations and 601,696,580 for answer generation.

\section{Conclusion}
MultiQ\&A is a crowdsourcing-based method to assess the robustness and consistency of LLM-generated answers. By perturbing 376,201 questions into 1,881,005 lexical variations while preserving semantics, our experiments across 13 datasets quantified consistency, reliability, and robustness, providing valuable insights into LLM responses. We believe that MultiQ\&A provides a promising infrastructure for institutions adopting LLMs with increased confidence.



\bibliography{aaai25}

\appendix

\section{Disclaimer}
{
This paper was prepared for informational purposes by the Artificial Intelligence Research group of JPMorgan Chase \& Co. and its affiliates ("JPMorgan'') and is not a product of the Research Department of JPMorgan. JPMorgan makes no representation and warranty whatsoever and disclaims all liability, for the completeness, accuracy or reliability of the information contained herein. This document is not intended as investment research or investment advice, or a recommendation, offer or solicitation for the purchase or sale of any security, financial instrument, financial product or service, or to be used in any way for evaluating the merits of participating in any transaction, and shall not constitute a solicitation under any jurisdiction or to any person, if such solicitation under such jurisdiction or to such person would be unlawful.
}

\end{document}